\documentclass[11pt,a4paper]{article}
\usepackage[hyperref]{emnlp2020}
\usepackage{times}
\usepackage{latexsym}
\usepackage{subcaption}

\usepackage{graphicx}
\usepackage{booktabs}

\usepackage{forest}
\usetikzlibrary{matrix}

\forestset{
    nice empty nodes/.style={
        for tree={
            s sep=0.1em, 
            l sep=0.33em,
            inner ysep=0.4em, 
            inner xsep=0.05em,
            l=0,
            calign=midpoint,
            fit=tight,
            where n children=0{
               tier=word,
               minimum height=1.25em,
            }{},
            where n children=2{
               l-=1em,
            }{},
            parent anchor=south,
            child anchor=north,
            delay={if content={}{
                    inner sep=0pt,
                    edge path={\noexpand\path [\forestoption{edge}] 
                    			(!u.parent anchor) 
                               -- (.south)\forestoption{edge label};}
                }{}}
        },
    },
}

\usepackage{tikz}
\newcommand*\circled[1]{\tikz[baseline=(char.base)]{
            \node[shape=circle,draw,inner sep=0.4pt] (char) {#1};}}

\usepackage{microtype}

\aclfinalcopy 

\title{Latent Tree Learning with Ordered Neurons: What Parses Does It Produce?}

\author{\bf Yian Zhang \\
Dept. of Computer Science\\
New York University\\
\href{mailto:yian.zhang@nyu.edu}{\tt yian.zhang@nyu.edu}
}
\date{}

\begin{document}
\maketitle
\begin{abstract}
Recent \emph{latent tree learning} models can learn constituency parsing without any exposure to human-annotated tree structures. One such model is ON-LSTM \citep{ONLSTMShen}, which is trained on language modelling and has near-state-of-the-art performance on unsupervised parsing. In order to better understand the 
performance and consistency of the model as well as how the parses it generates are different from gold-standard PTB parses, we replicate the model with different restarts and examine their parses. We find that (1) the model has reasonably consistent parsing behaviors across different restarts, (2) the model struggles with the internal structures of complex noun phrases, (3) the model has a tendency to overestimate the height of the split points right before verbs. We speculate that both problems could potentially be solved by adopting a different training task other than unidirectional language modelling.

\end{abstract}

\section{Introduction}
\emph{Grammar induction} is the task of learning the grammar of a target corpus without exposure to the parsing ground truth or any expert-labeled tree structures \citep{Charniak-AAAI,klein-manning-2002-generative}. Recently emerging \emph{latent tree learning} models provide a new approach to this problem \citep{DBLP:conf/iclr/YogatamaBDGL17,DBLP:journals/corr/MaillardCY17,Choi-Gumbel,PRPNShen,urnng}. They learn syntactic parsing under only indirect supervision from their main training tasks such as language modelling and natural language inference.

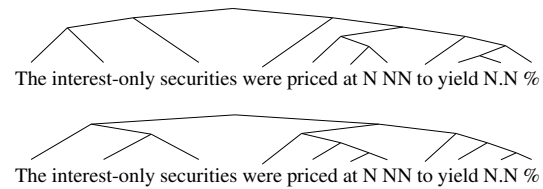
\begin{figure}[!t]

	\centering
	\hspace{0em}
	\vspace{0.5em}
	\scalebox{0.7}{
	\begin{forest}
		shape=coordinate,
		where n children=0{
			tier=word
		}{},
		nice empty nodes
        [ [ [ [The] [interest-only] ] [securities] ] [ [were] [ [ [priced] [ [at] [ N N\/N ] ] ] [ [to] [ [ [yield] [N.N] ] [\%] ] ] ] ] ]
	\end{forest}}
	\vspace{0.5em}
	\hspace{0em}
	\scalebox{0.7}{
	\begin{forest}
		shape=coordinate,
		where n children=0{
			tier=word
		}{},
		nice empty nodes
		[ [  [The] [[interest-only] [securities]] ] [ [ [were]  [ [priced] [ [at] [ N N\/N ] ] ] ] [ [to] [  [yield] [ [N.N]  [\%] ] ] ] ] ] ]
\end{forest}}
	
\caption{\label{fig:1} An example of an ON-LSTM's parse (top) disagreeing with a binary parse tree converted from a PTB gold-standard parse.
}
\end{figure}

In this study, we analyze ON-LSTM \citep{ONLSTMShen}, a new latent tree learning model that set the state of the art on unsupervised constituency parsing on WSJ test \citep{marcus-etal-1993-building} when it was published at ICLR 2019. The model is trained on language modelling and can generate binary constituency parsing trees of input sentences like the one in Figure \ref{fig:1}.
 
As far as we know, though there is an excellent theoretical analysis paper \citep{chris-dyer} of the ON-LSTM model that focuses on the model's architecture and its parsing algorithm, there is no
systematic analysis of the parses the model generates. There are no in-depth investigations of (i) whether the model's parsing behavior is consistent among different restarts or (ii) how the parses it produces are different from PTB gold standards. Answering these questions is crucial for a better understanding of the capability of the model and may bring insights into how to build more advanced latent tree learning models in the future.

Therefore, we replicate the model with 5 random restarts and look into the parses it generates. We find that (1) ON-LSTM has fairly consistent parsing behaviors across different restarts, achieving a self F1 of 65.7 on WSJ test. (2) The model struggles to correctly parse the internal structures of complex noun phrases. (3) The model has a consistent tendency to overestimate the height of the split points right before verbs or auxiliary verbs, leading to a major difference between its parses and the Penn Treebank gold-standard parses. We speculate that both problems can be explained by the training task, unidirectional language modelling, and thus we hypothesize that training a bidirectional model on a more syntax-related task like acceptability judgement might be a good choice for future latent tree learning models.

\section{Related Work}


ST-Gumbel \citep{Choi-Gumbel} and RL-SPINN \citep{DBLP:conf/iclr/YogatamaBDGL17} are two earlier latent tree learning models. These models are designed to learn to parse input sentences in order to help solve a downstream sentence understanding task such as natural language inference. Since they are not designed to approximate PTB grammar \citep{marcus-etal-1993-building}, their unsupervised parsing F1's on WSJ test are relatively low (20.1 and 25.0).

PRPN \citep{PRPNShen} and URNNG \citep{urnng} are two of the stronger latent tree learning models that have comparable unsupervised parsing performance (F1=42.8 and 52.4) with ON-LSTM (F1=49.4). URNNG is based on Recurrent Neural Network Grammar \citep{rnng}, a probablitic generative model; PRPN is a neural language model that implicitly models syntax using a structured attention mechanism. 

\citet{williams2018} analyze ST-Gumbel and RL-SPINN. They find that though the two models perform well on sentence understanding, neither of the models induces consistent and non-trivial grammars.


Prior to this work, \citet{chris-dyer} also analyze ON-LSTM. They raise doubts on the necessity of the model's novel gates and mathematically prove that it is impossible for the parsing algorithm used by \citet{ONLSTMShen} to correctly parse a certain class of structures. In comparison, this study takes a more empirical approach that is similar to that of \citet{williams2018}.

\section{Data and Model}

\paragraph{WSJ Dataset}
WSJ is the Wall Street Journal Section of PTB \citep{marcus-etal-1993-building}, which is the most commonly used dataset for training and evaluating parsers including latent tree learning models \citep{williams2018,phu2018}. It is also the dataset ON-LSTM is originally trained on. We follow the traditional split of WSJ: sections 0-21 as WSJ train, section 22 as WSJ dev, and section 23 as WSJ test. We also use WSJ 10, a subset of WSJ that includes all sentences with length $<10$. In the experiments, the model is always trained on WSJ train on language modelling, and evaluated on WSJ test, WSJ dev and/or WSJ 10 on constituency parsing.

\paragraph{Models}
\begin{figure*}[ht]

\centering
\includegraphics[width=\textwidth]{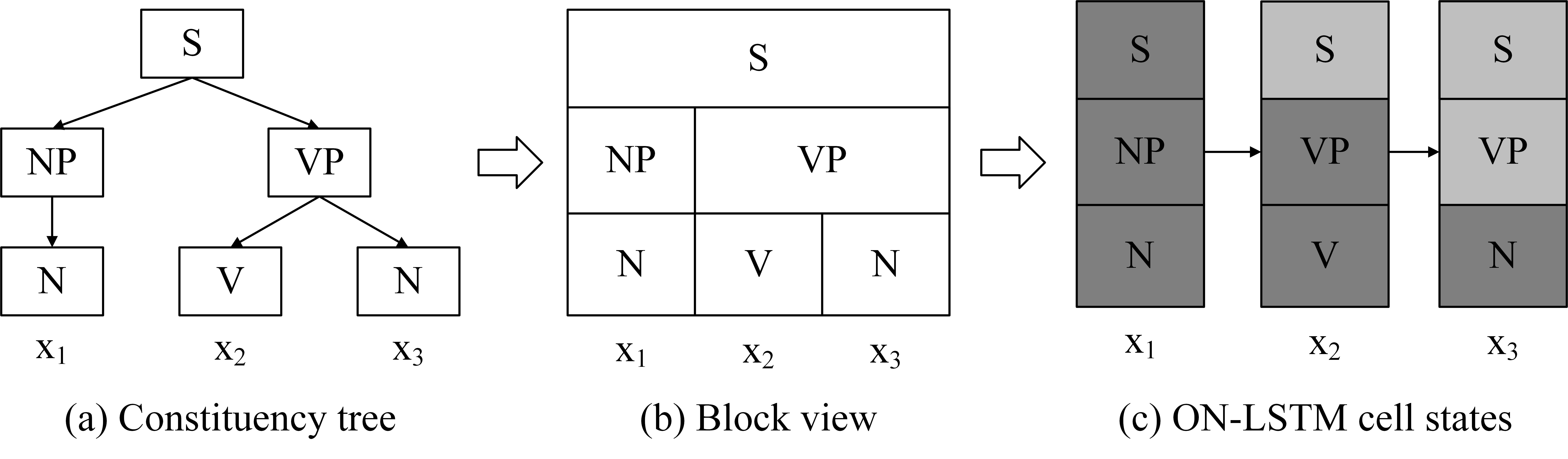}
\caption{\label{intuition} An example of the correspondences between a constituency parse tree and the hidden states of ON-LSTM. Intuitively, when performing language modelling, the information related to the highest-level constituent ``S'' is useful when predicting both token $x_2$ and token $x_3$, while the information related to the first ``N'' is only useful in predicting $x_2$, and can be erased after the prediction of $x_2$. In order to avoid removing ``S'' information when removing ``N'' information, the model will store ``S'' information in higher dimensions, and information of ``N'' in lower dimensions. Image source: \citet{ONLSTMShen}}
\end{figure*}

ON-LSTM is an LSTM model \citep{HochSchm97} plus a novel activation function, which causes the model to learn to store long-term information in high-ranking dimensions, implicitly encoding a constituency parse. The model is equipped with a master forget gate $\widetilde{f_t}$ and a master input gate $\widetilde{i_t}$. At each timestep, $\widetilde{f_t}$ is multiplied element-wise to the previous cell state $c_{t-1}$ and thus controls to what extent the value of each dimension in the previous cell state can be forgotten; $\widetilde{i_t}$ is multiplied element-wise to the candidate update values $\hat{c_t}$ and thus controls how much new information can be written to each dimension in the cell state. The values of $\widetilde{f_t}$ and $\widetilde{i_t}$ are computed at each timestep based on the input token and the cell state. The model uses $cumax()$ as the activation function of the master gates, where $$cumax(*):=cumsum(softmax(*))$$ $$cumsum(\vec{a}):=[a_1,a_1+a_2,..., \sum_{i=1}^k a_i, ..., \sum_{i=1}^n a_i] $$ Therefore, the values in $\widetilde{f_t}$ are always monotonically increasing from 0 to 1, and the values in $\widetilde{i_t}$ are always monotonically decreasing from 1 to 0. As a result, when a dimension is updated/erased, all of the dimensions whose ranks are lower than it are also updated/erased. Intuitively, in an extreme and simplified example where $\widetilde{f_t}=(0, ..., 0, 1, ..., 1)$, the model is just picking a dimension $d$, erases all the dimensions from $1$ to $d-1$ in $c_{t-1}$, and keeps dimensions $>d$ unchanged.



As a result of this novel updating rule, the model will tend to store long-term information in high-ranking dimensions and short-term information in low-ranking dimensions so that when the model frequently erases and updates low-ranking dimensions of the cell state, the long-term information stored in high-ranking dimensions will stay unaffected. When the model is trained to perform language modelling, since a higher-level constituent always spans more words than its children, its related information will continuously be useful for word prediction in a longer term and will thus be stored in higher-ranking dimensions (see Fig \ref{intuition} for an example). Therefore, intuitively, if a high-ranking dimension is erased/updated, it probably means that the currently processed input token is the start of a new high-level constituent.

Based on this intuition, the master forget gates can be used to perform binary constituency parsing. In binary parsing, a constituent (which is initially the whole sentence) is recursively split into two constituents until each constituent contains only one word. Therefore, each space between each pair of adjacent words is a split point the parsing algorithm will use at some point to make a split, and the order in which these split points are used decides what the resultant parsing tree will be like. In the case of ON-LSTM, the parsing algorithm uses the split points in the decreasing order of their ``height", where the height of a split point between $x_{t-1}$ and $x_t$ is usually defined as $\hat{d}^f_t$\footnote{As an exception, the height of the split point between $x_1$ and $x_2$ is defined as max($\hat{d}^f_1$, $\hat{d}^f_2$).}, an estimate of the transition point\footnote{\citet{ONLSTMShen} use the term ``split point". We use a different term to avoid confusion with the more frequently used ``split point" concept in this paper, which means the space between two words.} in $\widetilde{f_t}$ from the 0-segment\footnote{The master forget gate computed using the cumax() function is an expectation of a binary gate g=(0, ..., 0, 1, ..., 1), and the rank of the first ``1'' indicates to what extent the currently processed input word contains high-level information. For formal mathematical expressions of the model architecture, we encourage you to read the original model paper.} where values are small and close to 0 to the 1-segment where values are large and close to 1. Intuitively, the more information is forgotten at a timestep, the higher the split point before the token.





We train ON-LSTM with 5 different random seeds on WSJ train using hyperparameters shared by \citet{ONLSTMShen}. Note that the training objective is language modelling, so we only use the sentences from WSJ train and the model never has access to the parsing trees in the dataset or any other tree structures. On WSJ test, our models achieve an average perplexity of 56.33 ($\pm$0.06) on language modelling and average F1 score of 46.43 ($\pm$1.79) on unsupervised parsing, while the original paper reports 56.17 ($\pm$0.12) and 47.7 ($\pm$1.5). This shows that we roughly reproduce their work\footnote{The 5 ONLSTM models we train and their parses can be found at \href{https://github.com/YianZhang/ONLSTM-analysis}{\url{https://github.com/YianZhang/ONLSTM-analysis}}}.

\section{Experiments}
 To analyze the model's consistency, we use the 5 models we train to parse WSJ test and WSJ 10, calculate the self F1 and standard deviation on each dataset, and compare them to that of the random baseline. Self F1 is the average of unlabeled binary F1 scores between every pairing of the five parses, each produced by one model. It shows to what extent each model agrees with the parsing decisions of the other four. 
 
 To take a closer look at the parses generated by the model, we then use our 5 models to parse WSJ dev, and report the average of the models' parsing accuracies on each constituent type. We use the constituent-level accuracy as a guide to analyze how the parses ON-LSTM produces are different from PTB gold standards.

In the experiments, we include a simple random baseline that produces parses by recursively and randomly splitting the sequences to two halves. This is the same with ON-LSTM's parsing algorithm except that the baseline model chooses split points in a random order.
 
 \section{Results}
 
 \begin{table}[t]
\small \centering
\begin{tabular}{lcccc}
\toprule
\bf Model & \bf Test Set & \multicolumn{2}{c}{\bf F1($\sigma$)} & \bf Self F1 \\

\midrule
ON-LSTM layer1 & WSJ test & 23.4 & 2.1 & 56.5 \\
\bf ON-LSTM layer2 & \bf WSJ test & \bf 46.4 & \bf 1.8 & \bf 65.7 \\
ON-LSTM layer3 & WSJ test & 31.7 & 7.5 & 38.1 \\

ON-LSTM layer1 & WSJ10 & 44.4 & 3.0 & 71.6 \\
\bf ON-LSTM layer2 & \bf WSJ10 & \bf 69.8 & \bf 1.9 &\bf 82.1 \\
ON-LSTM layer3 & WSJ10 & 54.1 & 8.5 & 56.6 \\
\midrule
Random & WSJ test & 20.3 & 0.1 & 24.8 \\ 
Random & WSJ10 & 39.3 & 0.1 & 40.7 \\

\bottomrule
\end{tabular}  

\caption{\label{tab:selff1} F1, standard deviation, and self F1 of different layers of ON-LSTM on WSJ test and WSJ 10. }
\end{table}

\subsection{Does the model learn consistent grammars?}
The self F1's of ON-LSTM are shown in Table \ref{tab:selff1}. On both datasets, all three layers of ON-LSTM show much higher self F1 than the random baseline. This shows the model produces fairly consistent parses across different restarts. The 2nd layer, the layer with the highest parsing F1, is also the most consistent layer according to its self F1 and standard deviation. Its self F1 scores on both WSJ test and WSJ 10 are $\sim41$ higher than that of the random baseline.

\subsection{How are ON-LSTM's parses different from PTB parses?}

In this experiment, we focus on layer 2 of the model. For each model restart, we compute its parsing accuracy of every non-unary constituent type that occurs $>5$ times in WSJ dev (sentence-level occurrences aside, explained below).  We average the accuracies of the 5 restarts and list the results in Table \ref{tab:accuracy}. The constituent types are listed in decreasing order based on the difference between ON-LSTM accuracy and the random baseline accuracy.

Different from the previous works \citep{williams2018,ONLSTMShen,phu2018}, we do not take into account any constituent that spans over an entire sentence, because any parser has 100\% accuracy on these constituents. The way we compute the accuracy better reveals the model's command of each constituent type, and makes comparisons across constituent types more fair, since some types are more likely to appear as full sentences. We follow the clues in the accuracies to look into the parses generated by the models and find two cases where the models struggle, as we discuss in the following sections.

\begin{table}[t!]
\small \centering
\scalebox{1}{
\begin{tabular}{lccccc}
\toprule
\bf Constituent & \multicolumn{2}{c}{\bf Accuracy($\sigma$)} & \multicolumn{2}{c}{\bf Random($\sigma$)} & \bf $\Delta$Acc \\

\midrule
 
SQ  & 77.8 & 0.0 & 15.6 & 15.1 & 62.2 \\
VP  & 55.5 & 2.2 & 12.5 & 0.3 & 43.0 \\
NP  & 56.8 & 5.5 & 22.5 & 0.4 & 34.3 \\
PP  & 52.8 & 1.5 & 18.7 & 0.5 & 34.1 \\
WHNP  & 42.9 & 6.7 & 16.2 & 8.3 & 26.7 \\
S  & 32.6 & 3.6 & 10.8 & 1.0 & 21.8 \\
ADJP  & 44.9 & 7.9 & 24.5 & 1.7 & 20.3 \\
ADVP  & 44.8 & 4.2 & 26.2 & 3.2 & 18.6 \\
UCP  & 29.5 & 7.1 & 15.8 & 5.8 & 13.7 \\
QP  & 41.5 & 7.3 & 30.1 & 2.2 & 11.4 \\
SBAR  & 20.8 & 8.5 & 10.2 & 0.9 & 10.5 \\
NX  & 26.9 & 12.1 & 17.1 & 6.3 & 9.7 \\
SINV  & 11.7 & 11.3 & 8.3 & 7.5 & 3.4 \\
PRN  & 20.5 & 3.0 & 18.4 & 3.6 & 2.1 \\
WHPP  & 14.5 & 9.3 & 29.1 & 13.4 & -14.5 \\
NAC  & 5.9 & 5.0 & 21.5 & 2.8 & -15.6 \\
CONJP  & 15.0 & 5.0 & 37.5 & 11.2 & -22.5 \\
 \bottomrule
\end{tabular}}  

\caption{\label{tab:accuracy} ON-LSTM (layer 2)'s average parsing accuracies of non-unary constituents in WSJ dev across 5 restarts. The last column is the difference between the second and the fourth column.}
\end{table}

\paragraph{Complex Noun Phrases}
\label{long noun groups}
As shown in table \ref{tab:accuracy}, the model has a poor parsing performance on NX ($\Delta$acc=$9.7$) and NAC ($\Delta$acc=$-15.6$), in contrast to the good performance on NP ($\Delta$acc=$34.3$). NX and NAC are marker constituents that split an NP into smaller chunks. NX marks individual conjuncts in an NP, e.g. (NP the (NX (NX white shirt) and (NX blue jeans))). NAC shows the scope of a modifier within an NP, e.g. (NP (NAC Secretary (of (State))) James Baker). This contrast suggests that the model is able to identify noun phrases in a sentence, but fails to understand their internal structures. We inspect the model's parses of noun phrases that contain NX and NAC and find that the way the model splits these phrases is very random. We do not identify any pattern.

One can possibly attribute this failure to the use of language modelling as the training task. Whether ON-LSTM makes a split at a token depends on how much information the model chooses to forget at this timestep. Since the model is trained for unidirectional language modelling, it decides whether to forget certain information based on whether the information will be helpful for word predictions in the future. However, constituents inside the same complex noun phrase are sometimes closely related, and cross-constituent hints can be helpful to word predictions. In the NX example we give,  ``white shirt'' gives important hints for the model to predict the tokens ``blue jeans'', as it suggests that the tokens after ``and'' might be a color followed by a type of clothing. This may be why the model might choose not to forget much information after ``white shirt and'', leading to a missing split between ``and" and ``blue". 

\paragraph{Split Points Right Before Verbs}

\label{verbs}
As shown in Table \ref{tab:accuracy}, the model's parsing performances on SQ and VP are the best ($\Delta$acc=$62.2$ and $43.0$, ranking the first and second among all constituents), while it does not parse SBAR (subordinate clauses) in a way that is similar to PTB parses ($\Delta$acc=$10.5$, Acc=20.5). Based on this clue, we look into the parses and find the model has a consistent and strong tendency to overestimate the height of the split point right before a verb.

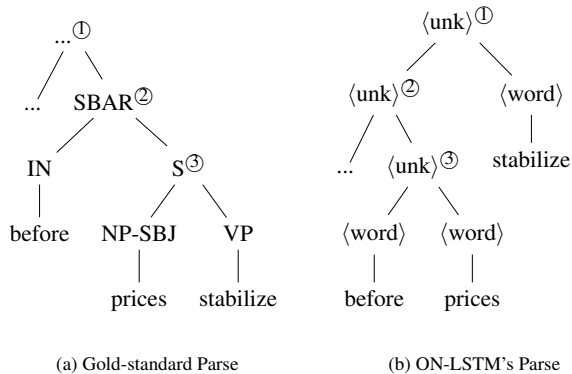
\begin{figure}[t]
\scalebox{0.8}{
	\centering
\begin{subfigure}[b]{0.3\textwidth}
	\begin{forest}
	[...\textsuperscript{\small{\textcircled{1}}}
	[...]
	[SBAR\textsuperscript{\small{\textcircled{2}}}
	[IN[before]]
	[S\textsuperscript{\small{\textcircled{3}}}
	[NP-SBJ[prices]]
	[VP[stabilize]]
	]
	]
	]
    \end{forest}
    \vspace{1.0 em}
    \caption{Gold-standard Parse}
\end{subfigure}
\hspace{1.0 em}
\begin{subfigure}[b]{0.3\textwidth}
	\begin{forest}
	[$\langle$unk$\rangle$\textsuperscript{\small{\textcircled{1}}}
	[$\langle$unk$\rangle$\textsuperscript{\small{\textcircled{2}}}
	[... ]
	[$\langle$unk$\rangle$\textsuperscript{\small{\textcircled{3}}}
	[$\langle$word$\rangle$ [before]]
	[$\langle$word$\rangle$[prices]]
	]
	]
	[$\langle$word$\rangle$ [stabilize]]
	]
\end{forest}
\vspace{1.0 em}
\caption{ON-LSTM's Parse}
\end{subfigure}
}
\caption{\label{fig:trees} An example of ON-LSTM overestimating the height of the split point right before the verb. (a) is the gold-standard parsing tree; (b) is the binary tree produced by ON-LSTM. $\circled{1}$, $\circled{2}$, and $\circled{3}$ mark the order/height of the split points in each parse.
}
\end{figure}

We inspect the model's parses of sentences that contain subordinate clauses, and find that a common mistake made by ON-LSTM is to assign a higher height to the split point right before the main verb of the clause than to the split points right before/after the start/end of the clause. Since ON-LSTM parses a sentence by recursively splitting the sentence at the highest split point, this means the subordinate clause will show up separately in two different constituents rather than a complete single constituent in the parse generated by ON-LSTM. For example, as shown in figure \ref{fig:trees}, split point \circled{1} of a gold-standard parser is right \textbf{before} the token ``before'' and it splits the upper constituents into two parts:  ``...'' and SBAR, where SBAR contains exactly three words: ``before'', ``prices'', and ``stabilize''. In contrast, ON-LSTM chooses the split point right before the verb ``stabilize'' as split point \circled{1} and thus in its parse there is no constituent that contains exactly these three words.

According to our observations, this behavior is not incidental. We randomly sample 30 SBARs from WSJ dev. For each SBAR, we observe whether the first (highest) split point inside the clause (border tokens included) chosen by each model is (1) right before the main verb/auxiliary verb, (2) right before the first token or right after the last token of the clause, or (3) the other tokens in the clause. For example, Figure \ref{fig:trees} (a) is of case (2) and Figure \ref{fig:trees} (b) is of case (1).  We compute the percentages of case (1) and case (2) for each ON-LSTM model and show them in Table \ref{tab:proof}. We find that all 5 models have a much stronger tendency than the gold-standard parser to choose the split point right before the verb as the highest split point inside a subordinate clause. On each row, the two numbers add to nearly 100, which means when the model makes a mistake on SBAR, it is almost always because it makes the highest split right before the verb.


\begin{table}[t]
\centering
\scalebox{0.85}{
 \begin{tabular}{| c | c | c |} 
 \hline
 Model & Case 1: Verb (\%) & Case 2: Border (\%)\\
 \hline
 ON-LSTM 1 & 90.0 & 10.0  \\ 
 \hline
 ON-LSTM 2 & 70.0 & 20.0  \\
 \hline
 ON-LSTM 3 & 70.0 & 16.7 \\
 \hline
 ON-LSTM 4 & 73.3 & 16.7 \\
 \hline
 ON-LSTM 5 & 66.7 & 30.0 \\
 \hline
 Gold & 0.0 & 100.0 \\
 \hline
\end{tabular}}
\caption{\label{tab:proof} The percentage of times the split point right before the main verb and the split point right before/after the border tokens in a subordinate clause is the highest split point in the clause. The last row represents the gold standard.}
\end{table}

This tendency also explains why the model's parsing accuracy is the highest on VP and SQ, two constituents which almost always start with a verb. As discussed earlier in this section, a constituent will be correctly parsed if and only if no split point inside it is higher than the split points right before/after the start/end token. Therefore, constituents starting with a verb are naturally easier for ON-LSTM because of this tendency.

A possible reason of this tendency is that since the model is trained on unidirectional language modelling, when it predicts the height of a split point before a token, it only has access to the current token and all the tokens before it. However, when the current input token is a beginning word of a subordinate clause such as ``as'', ``which'', ``after'', it is usually impossible to tell whether it is the start of a subordinate clause. Counterexamples are ``as soon as possible'', ``which to choose'', ``after 2 hours'', etc. Meanwhile, the model probably learns that the appearance of a verb almost always means the start of a high-level constituent VP. As a result, it assigns high heights to split points right before verbs and ignores higher-level constituents including SBAR. If this is true, then a natural and direct fix of this problem is to adopt a bidirectional task such as masked language modelling instead.

\section{Conclusions and Future Work}
In summary, the model shows basic self-consistency on the task of constituency parsing, and it is consistently able to correctly identify certain constituents (SQ, VP, NP). All these results show that the unique design of the model brings us closer to developing consistently powerful unsupervised parsing models. However, the experiments show that it (a) struggles with the internal structures of complex NPs, and (b) often overestimates the height of the split points right before verbs. Based on our analysis, we hypothesize that both of the failures can be at least partially attributed to the use of unidirectional language modelling as the training task. 

There are two potential problems with this training task. First, the motivation of language modelling generally does not perfectly match the target task constituency parsing, since cross-constituent hints are sometimes helpful, as revealed by (a). Second, it is very hard for a unidirectional model to correctly identify some high-level constituents, as revealed by (b). Therefore, we believe a promising research direction is to build latent tree learning models based on bidirectional model architectures like transformer \citep{transformers} and the task of acceptability judgement with a dataset like CoLA \citep{cola}, which is a more syntax-related sentence-level task that requires the model to predict whether an input sentence is grammatically acceptable. Another option to consider is masked language modelling because it is also a bidirectional task and is much easier to scale up compared to acceptability judgement since it is a self-supervised task.

\section*{Acknowledgments}
We appreciate Sam Bowman for giving valuable overall project feedbacks and suggestions; we appreciate Phu Mon Htut for patiently sharing and explaining the code and experiment details of her study; we appreciate Yikang Shen for making their code public and granting us the right to reuse the figures in their paper. We would also like to thank Alex Warstadt and Daniel Chin for their great writing suggestions.

\bibliographystyle{acl_natbib}
\bibliography{anthology,emnlp2020}

\end{document}